\documentclass[runningheads]{llncs}

\usepackage{eccv}

\usepackage{eccvabbrv}

\usepackage{graphicx}
\usepackage{booktabs}

\usepackage[accsupp]{axessibility}  

\usepackage{hyperref}

\usepackage{orcidlink}
\usepackage[linesnumbered,ruled,vlined]{algorithm2e} 

\usepackage{mathrsfs}   
\usepackage{multirow}
\usepackage{booktabs}
\usepackage{tabularx}
\usepackage{floatrow}
\usepackage{colortbl} 
\usepackage{xcolor} 

\makeatletter
\newcommand{\removelatexerror}{\let\@latex@error\@gobble}
\makeatother

\newcolumntype{Y}{>{\centering\arraybackslash}X} 

\newcommand{\tr}{} 
\newcommand{\ddf}{\textit{ddf}} 
\newcommand{\csection}[1]{\vspace{0.5em} \noindent \textbf{#1}} 

\begin{document}

\title{UL-VIO:\\ Ultra-lightweight Visual-Inertial Odometry \\ with Noise Robust Test-time Adaptation}

\titlerunning{UL-VIO: Ultra-lightweight Visual-Inertial Odometry}

\author{Jinho Park\inst{1}\orcidlink{0000-0003-1613-0769} \and
Se Young Chun\inst{2}\orcidlink{0000-0001-8739-8960} \and
Mingoo Seok\inst{1}\orcidlink{0000-0002-9722-0979}}

\authorrunning{J.~Park et al.}

\institute{Columbia University, New York NY 10027, USA \and
Dept. of ECE, INMC \& IPAI, Seoul National University, Republic of Korea\\
\email{jp4327@columbia.edu,~sychun@snu.ac.kr,~ms4415@columbia.edu}}

\maketitle
\begin{abstract}
Data-driven visual-inertial odometry (VIO) has received highlights for its performance since VIOs are a crucial compartment in autonomous robots.
However, their deployment on resource-constrained devices is non-trivial since large network parameters should be accommodated in the device memory.
Furthermore, these networks may risk failure post-deployment due to environmental distribution shifts at test time.
In light of this, we propose \textbf{UL-VIO} -- an ultra-lightweight ($<1$M) VIO network capable of test-time adaptation (TTA) based on visual-inertial consistency.
Specifically, we perform model compression to the network while preserving the low-level encoder part, including all BatchNorm parameters for resource-efficient test-time adaptation.
It achieves $36 \times$ smaller network size than state-of-the-art with a minute increase in error -- $1$\% on the KITTI dataset.
For test-time adaptation, we propose to use the inertia-referred network outputs as pseudo labels and update the BatchNorm parameter for lightweight yet effective adaptation.
To the best of our knowledge, this is the first work to perform noise-robust TTA on VIO.
Experimental results on the KITTI, EuRoC, and Marulan datasets demonstrate the effectiveness of our resource-efficient adaptation method under diverse TTA scenarios with dynamic domain shifts.


\end{abstract}    
\section{Introduction}
\label{sec:intro}

\begin{figure}[t]
    \centering
    \includegraphics[width=1\linewidth]{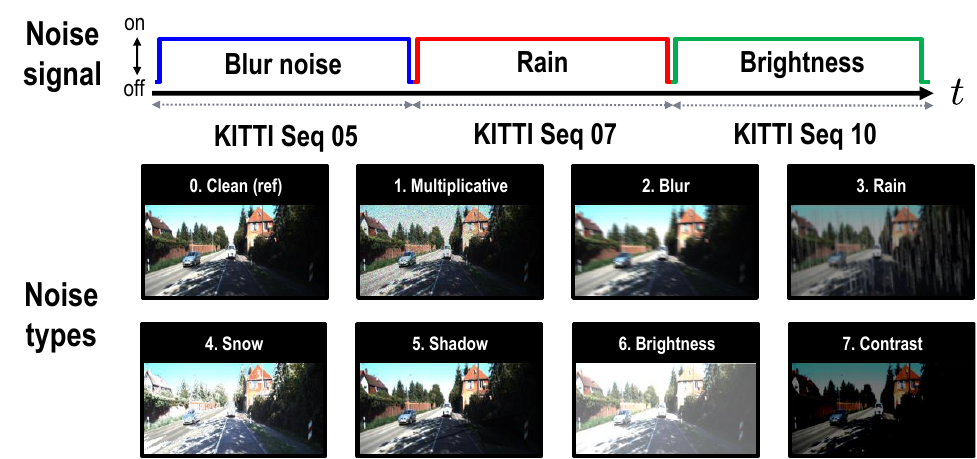}
    \caption{We address a domain shift problem that is likely to occur during driving scenarios. To emulate real-world driving scenarios, we introduce various vision noises into the image sequence inputted into the VIO model. We continuously run multiple odometry sequences to assess test-time adaptation without forgetting.}
    \label{fig:graph_abst}
\end{figure}

Deep learning-based visual-inertial odometry (VIO) \cite{chen2023search, yang2022efficient} has surpassed the performance of state-of-the-art geometry-based methods such as ORB-SLAM \cite{mur2015orb}.
Estimating one's ego-motion from camera images and inertial measurement unit (IMU) data sequences \cite{mur2015orb, forster2016manifold}, VIO is a crucial component in the autonomous navigation pipeline \cite{wang2017deepvo,teed2021droid,yang2018deep}.
However, deploying these networks on mobile autonomous platforms poses a significant challenge due to the limited memory and computing capacity of such devices.
More importantly, accessing off-chip DRAM memory requires two to three orders of magnitude more power compared to on-chip memory access \cite{wulf1995hitting, horowitz20141}, thereby imposing a significant limitation on the size of the networks that can be deployed on these platforms.
Although reducing the computational complexity of VIO has been studied in \cite{cai2020vonas, chen2023search}, their model size being over a couple of $10$ M hinders edge deployment.
In view of this, we target a model with $<1$M parameters to be entirely hosted within a tight on-chip memory in mobile hardware.

Yet another concern for mobile VIO platforms is that they may suffer from post-deployment performance degradation when encountering out-of-distribution (OoD) data at test time.
For example, a network trained on clean camera image sequences might be prone to failure when the image distribution shifts due to environmental conditions, e.g., shadow, snow, and rain.
To the best of our knowledge, none of the prior arts have investigated noise-robust test-time adaptation for VIO although train-time augmentation for noise-robustness was explored in \cite{chen2019selective}.
This motivates us to consider the effect of visual noise in VIO systems.
As shown in Fig. \ref{fig:graph_abst} during the video sequence, the network receives image streams of unseen distribution that differ from the source domain.

To ameliorate distribution shifts in classification tasks, researchers have proposed test-time adaptation (TTA) to modify the network on OoD downstream tasks \cite{hendrycks2019robustness, wang2021tent, wang2022continual, yuan2023robust, song2023ecotta, song2023test, boudiaf2022parameter,yuan2023robust, chen2022contrastive, zhang2023domainadaptor, chen2023improved, hatem2023point}.
However, conventional methods usually target image classification or semantic segmentation tasks that minimize prediction entropy at test time \cite{wang2021tent}.
Nevertheless, VIOs performing regression tasks cannot directly adopt such entropy-based methods simply due to a lack of prediction entropy.
Another way to utilize unlabeled data at test time is to spare a separate teacher network to generate pseudo labels \cite{wang2022continual}.
In autonomous ground and aerial vehicles, deploying a dedicated teacher network might not be feasible due to the large model size of a teacher network.
Hosting a teacher network in a remote server is also difficult because of the long latency.

To that end, we propose our resource-efficient test-time adaptation scheme based on multi-modal consistency loss.
Although inertia information is less precise than the visual one when no visual noise is present, it can be a relatively reliable sensor source under severe conditions \cite{yang2022efficient, chen2019selective}.
In light of this, our proposed TTA uses alternate modality-based prediction as the pseudo label can reduce the pose estimation error.
The contribution of our work is three-fold:
\begin{itemize}
    \setlength\itemsep{.5em}  
    \item We propose an ultra-lightweight visual-inertial odometry network with less than 1M parameters while keeping the low-level encoder part intact, including all BatchNorm (BN) parameters, to enable noise-robust test-time adaptation. It yields $36\times$ smaller model size than the state-of-the-art methods with comparable performance -- $1\%$ increase in pose estimation error.
    \item We introduce a resource-effective online adaptation for VIO using multimodal information in adverse conditions, efficiently handling quick transitions with only $5\%$ parameter overhead for inertial output. 
    \item Our proposed method was evaluated on the KITTI, EuRoC, and Marulan datasets with various vision corruptions. Under dynamic noise shifts, our model achieves up to $45\%$ reduction in translation RMSE ($18\%$ on average) through adaptation based on the KITTI dataset. 
\end{itemize}

\section{Related works}
\label{sec:related}

\subsection{Visual inertial odometry}
In recent years, end-to-end learning-based visual and visual-inertial odometry (VO, VIO) methods have gained interest owing to their performance in localization tasks \cite{chen2023search,yang2022efficient}.
VIO systems can continuously estimate an agent's ego-motion from sensor inputs, especially vision and inertial measurement unit (IMU) streams \cite{scaramuzza2011visual}. 
Precise localization is a crucial compartment of autonomous driving, robotics, and augmented reality.

After the first end-to-end network-based pose estimation work has been proposed in \cite{kendall2015posenet}, the problem has been reformulated into a sequence-to-sequence learning problem with the addition of IMU readings \cite{clark2017vinet}.
To perform sensor fusion in VIOs, a na\"ive concatenation was performed between visual and inertial features \cite{chen2023search, yang2022efficient, clark2017vinet}, deterministic or stochastic re-weighting of the combined features was introduced in by Chen \etal \cite{chen2019selective}, and attention-based fusion was proposed in ATVIO \cite{liu2021atvio}.

In pursuit of reducing the computational complexity of the network, skipping vision inference was proposed in \cite{yang2022efficient}, and network architecture search (NAS)-based computational complexity reduction was performed in \cite{cai2020vonas, yang2022efficient}.
However, for mobile deployment, it is crucial to minimize not only the number of floating point operations but also the model size based on the on-chip memory of the platform.
Communicating data from/to the off-chip memory typically consumes two to three orders of magnitude larger energy than the on-chip memory \cite{wulf1995hitting, horowitz20141}.
Prior works concentrate only on computational complexity reduction, neglecting the memory consideration.
Hence, we focus on model compression.

\subsection{Test-time adaptation}

While deep neural networks perform successfully on target domains, their performance may fall short of expectations when we execute the model in a real-world setting \cite{hendrycks2019robustness}.
Generating labels for the data stream at test time is expensive and may not be feasible in some situations. 
To that end, test-time adaptation (TTA) has been developed to modify pre-trained networks based on unlabeled target samples without the source data.

Recently, several works have proposed TTA for classification tasks. 
A foundational work, TENT \cite{wang2021tent}, proposed modifying only a small portion of the network by minimizing the entropy.
Following it, CoTTA \cite{wang2022continual} attempts to make the model adapt to continually changing environments at the cost of updating the entire network based on a teacher network.
EcoTTA \cite{song2023ecotta} allows the model to be updated more efficiently using a meta-network.
Song \etal proposed a TTA method that can utilize previously learned knowledge by dynamically switching a portion of the model depending on the sub-target domain \cite{song2023test}. 
LAME \cite{boudiaf2022parameter} resolves hyperparameter sensitivity during TTA.

Though not directly related to noise-robust TTA, adaptation to dataset change is proposed in \cite{li2020self} by utilizing meta-learning \cite{finn2017model} and self-supervision \cite{zhou2017unsupervised}.
Unsupervised learning of pose estimation by using DepthNet was first suggested in SfMLearner \cite{zhou2017unsupervised} and GeoNet \cite{yin2018geonet}.
However, since the network inference solely relies on visual modality and adapts itself based on self-generated warped features, their robustness to noise may not be guaranteed.

On the other hand, XVO \cite{lai2023xvo} utilizes the teacher model and auxiliary tasks like audio prediction to perform semi-supervision.
Similarly, CoVIO \cite{vodisch2023covio} employs replays to make online adaptations for different datasets.
Using additional networks occupying tens of millions of parameters for self-supervision as employed in the above-mentioned works may not be amenable in mobile settings with limited memory and energy constraints.

\section{Methods}
\label{sec:methods}

\begin{figure*}[t]
    \centering
    \includegraphics[width=\linewidth]{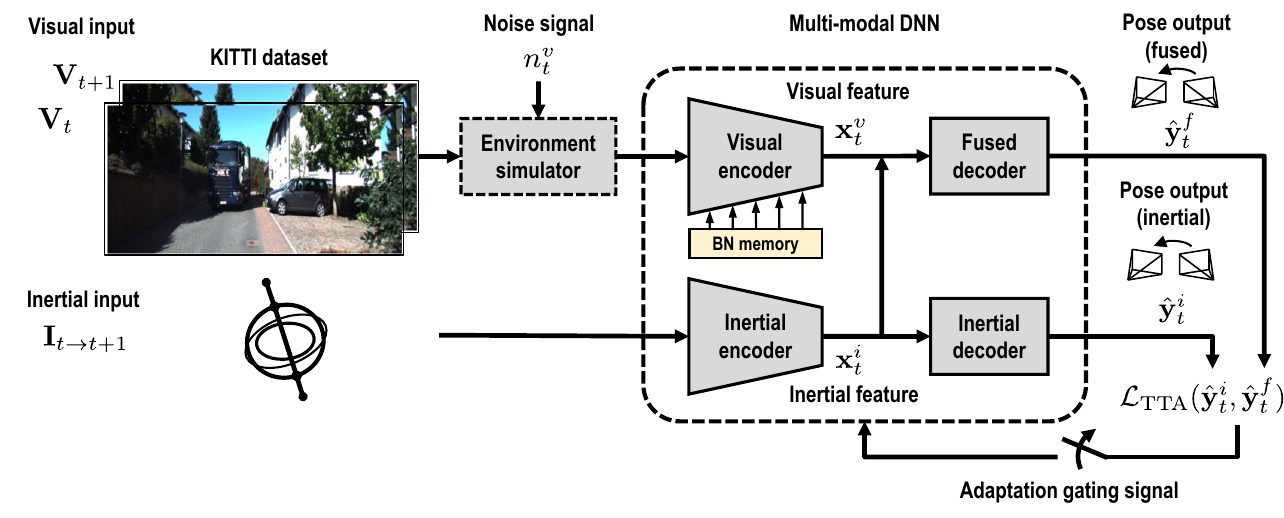}
    \caption{\textbf{Overall framework setup for UL-VIO.} The network has two input streams -- visual and inertial. Modulated by the noise signal, the environment simulator emulates the adversarial weather conditions. The network adapts using inertial input as the pseudo label when the adaptation gating signal is turned on. Parallel multi-modal encoders independently generate the visual and inertial features. Two pose outputs are generated based on visual-inertial feature fusion or inertial-only.}
    \label{fig:overall}
\end{figure*}

\begin{figure}[t]
    \centering
    \includegraphics[width=\linewidth]{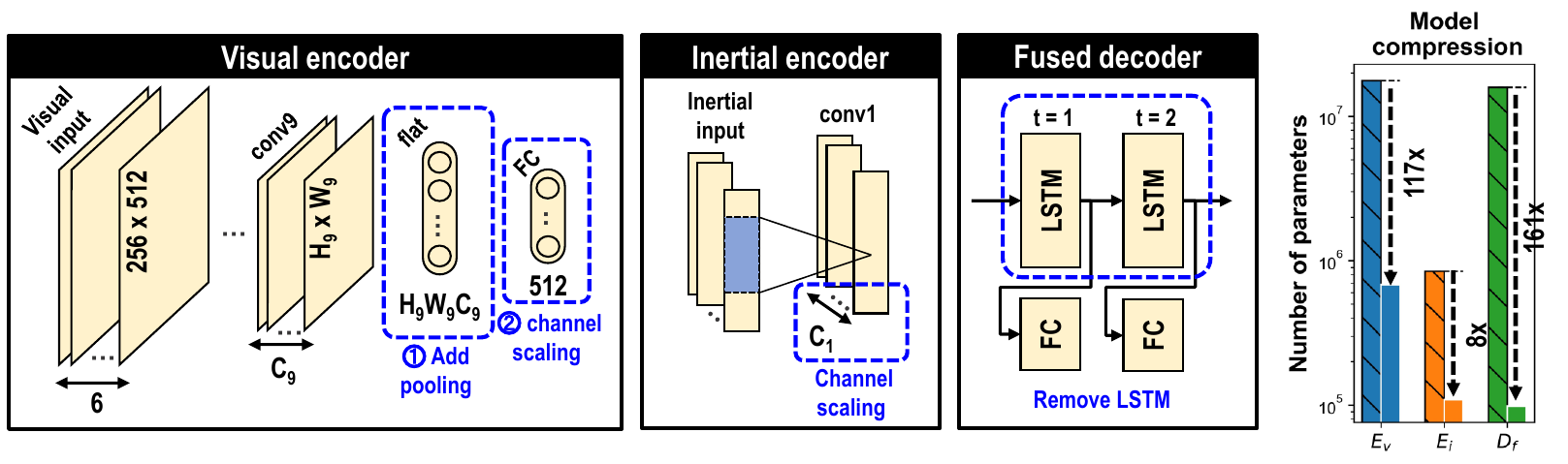}
    \caption{\textbf{Model compression.} We shrink the module size but keep the low-level parts in the visual encoder, including all BN parameters, to ensure test-time adaptation. We achieve $\{117 \times, 8 \times,161 \times\}$ reduction in $\{E_\text{visual}, E_\text{inertial},D_\text{inertial}\}$}
    \label{fig:model_comp}
\end{figure}

\subsection{Ultra-lightweight model compression}

\textbf{Network setup} Our pre-trained end-to-end VIO network deduces locomotion by inferring from visual and inertial data. 
It can also adapt to noisy visual inputs using multi-modal consistency when demanded.
As shown in Fig. \ref{fig:overall}, our VIO receives consecutive images $\{\mathbf{V}_i\}_{i=1}^N$ and $r$-times oversampled IMU data $\{\mathbf{I}_i\}_{i=1}^{Nr}$ as inputs. It then estimates a sequence of poses $ \{\mathbf{p}_t \}_{t=2}^T$ from the starting pose $\mathbf{p}_1$.
Here, $\mathbf{V}_i \in \mathbb{R}^{c \times h \times w}$, $\mathbf{I}_i \in \mathbb{R}^{6}$, and $\mathbf{p}_t \in \mathbf{SE}(3)$.
Such a sequence of poses is associated with 6-DoF agent pose transformations $\mathbf{T}_{t \rightarrow t+1}$ defined by $\mathbf{p}_t \mathbf{T}_{t \rightarrow t+1} = \mathbf{p}_{t+1}$. 
The transformation $\mathbf{T}_{t \rightarrow t+1}$ can be decomposed into a rotational component $\mathbf{\Phi}_t \in \mathbb{R}^3$ and a translational component $\mathbf{v}_t \in \mathbb{R}^3$.

The learning-based VIO has two encoders and two decoders. 
Except for the additional inertial decoder for multi-modal inference, the network follows generic VIO networks \cite{yang2022efficient, chen2023search, chen2019selective}.
The visual feature encoder $E_{\text{visual}}$ and the inertial feature encoder $E_{\text{inertial}}$ independently outputs the visual feature $\mathbf{x}_t^v$ and the inertial feature $\mathbf{x}_t^i$ from consecutive image frames $\mathbf{V}_{t \rightarrow t+1}$ and inertial measurement streams $\mathbf{I}_{t \rightarrow t+1}$ as in
\begin{equation}
    \mathbf{x}_t^v = E_{\text{visual}}(\mathbf{V}_{t \rightarrow t+1}), ~~~
    \mathbf{x}_t^i = E_{\text{inertial}}(\mathbf{I}_{t \rightarrow t+1})
\end{equation}
These feature vectors are then used by the decoders to estimate the pose $\hat{\mathbf{y}}$:
\begin{equation}
    \hat{\mathbf{y}}_t^f = {D}_{\text{fused}}(\mathbf{x}_t^v \mathbin\Vert \mathbf{x}_t^i), ~~~
    \hat{\mathbf{y}}_t^i = {D}_{\text{inertial}}(\mathbf{x}_t^i)
\end{equation}
where $\Vert$ denotes the concatenation operation. 
The estimated pose can also be expressed as $\mathbf{y}_t = \mathbf{\Phi}_t \Vert \mathbf{v}_t$.

\csection{Model compression} We target sub-million parameter count for the model to be accommodated in the on-chip memory of a mobile platform.
Commercial mobile processors like Apple A16 and Qualcomm Snapdragon only possess a few MB of on-chip memory.
We reduce the size of the visual encoder while maintaining the BN parameter size for test-time adaptation since tuning BN is a preferred method for adaptation. 
We aggressively downsize the decoder since the decoder evaluates the pose from domain invariant features.
We perform model compression on the state-of-the-art NASVIO \cite{chen2023search}.
Although NASVIO effectively reduces the computational complexity through network architecture search (NAS), its parameter count remains high as its $\{{E}_\text{visual}$, ${E}_\text{inertial}$, ${D}_\text{fused}\}$ occupy $\{17.69 \text{ M}$, $0.85 \text{ M}$, $15.88 \text{ M}\}$ parameters.

We resolve the bottleneck posed by the output feature map of the last convolutional layer by adding a pooling layer to reduce the tensor size.
The structure of the visual encoder, especially the BN parameters, is maintained as these will be updated to support adaptation.
We also reduce the channel size since this quadratically decreases the parameter count in 1-D and 2-D convolutional layers.
Many of the weights in a network are usually dominated by the deeper layers since the channel size has progressively grown.
Moreover, we replace the long short-term memory (LSTM) with a fully connected (FC) layer since this can reduce the model size by about $4 \times$, assuming the same feature size.
While prior research has employed LSTM to leverage temporal relationships, we find an FC layer with orders of magnitude smaller parameter numbers as the decoder can perform comparably -- incurring only $1\%$ increase in pose error.

As shown in Fig. \ref{fig:model_comp}, we summarize our approach and its effects in the following.
\begin{itemize}
    \item Add an AveragePool after the last convolutional layer in $E_\text{visual}$. This gives us $117 \times$ reduction in $E_\text{visual}$.
    \item Reduce the channel size in $E_\text{inertial}$ since the parameter number is quadratically proportional to it, attaining $8 \times$ compression in $E_\text{inertial}$.
    \item Replace the LSTM with fully connected layers for the $D_\text{fused}$, resulting in $161 \times$ downsizing in $D_\text{fused}$.
\end{itemize}

\csection{Loss function}
We use mean squared error (MSE) loss to train the network:
\begin{equation}
\mathcal{L}_\text{train} = \frac{1}{B} \sum_{j=1}^{B}   \left( 
                                                                \left\lVert \mathbf{v}_j-\hat{\mathbf{v}}_j^f \right\rVert^2_2
                                                                + \alpha \left\lVert \mathbf{\Phi}_j-\hat{\mathbf{\Phi}}_j^f \right\rVert^2_2
                                                         \right)
\label{eq:loss_train}
\end{equation}
where $\mathbf{v},\mathbf{\Phi}$ are the ground truth translational, rotational vectors, $\hat{\mathbf{v}},\hat{\mathbf{\Phi}}$ are the predicted counterparts. 
Here, $\lVert \cdot \rVert_2$ denotes $l^2$ norm, $B$ is the batch size, and $\alpha$ is the weight between translational and rotational components.


\begin{figure}[t]
    \centering
    \includegraphics[width=\linewidth]{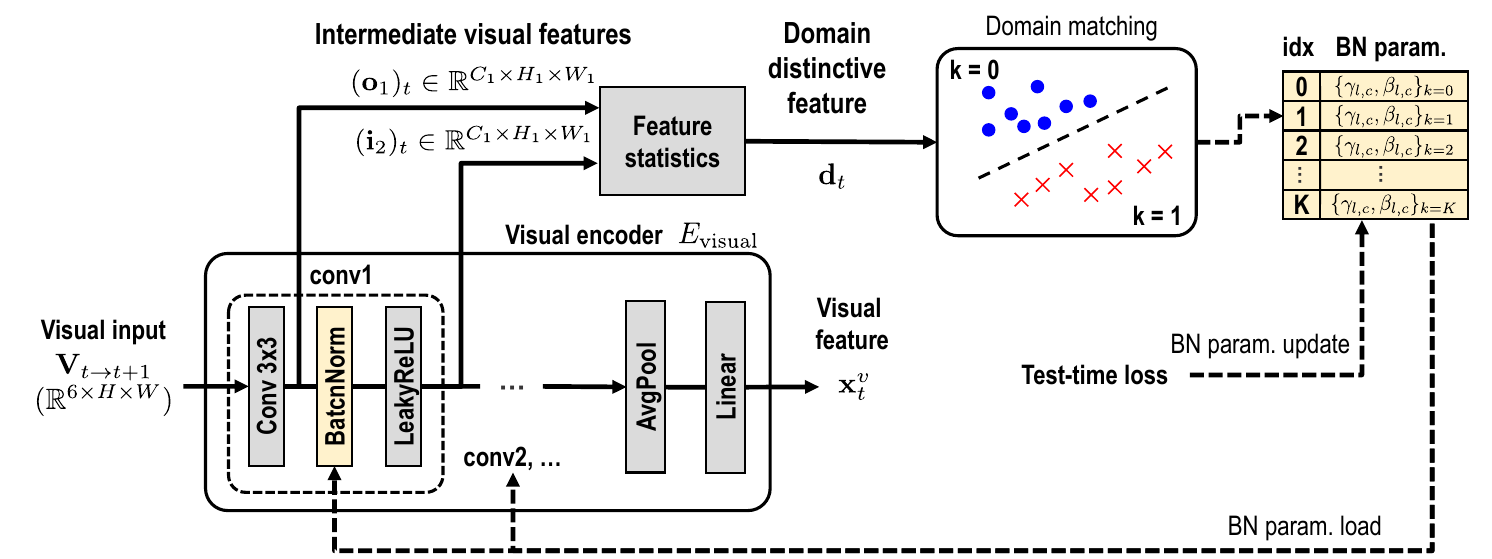}
    \caption{\textbf{Lightweight visual encoder with dictionary-based adaptation.} The statistics of intermediate feature maps during and after the first layer are taken to generate {\ddf}s. Although aggressively reducing the visual parameter footprint, we maintain the BN parameters intact for adaptation.}
    \label{fig:ve}
\end{figure}

\subsection{Test-time adaptation for lightweight VIO}
This section covers the visual encoder's noise detection and its adaptability.
Only the weights of the visual encoder are modified during adaptation while the weights of other modules are fixed.
As shown in Fig \ref{fig:ve}, domain distinctive features ({\ddf}s) from the early layers of the visual encoder are utilized for domain shift detection.
The visual encoder hosts an auxiliary dictionary to store and update learnable BN parameters corresponding to different noise types.
Domain shift detection and partial model updates have been studied in \cite{song2023test, park2023all}.

\begin{figure}[t]
 \removelatexerror
  \begin{algorithm}[H]
   
    \SetNlSty{normaltext}{}{:}
    \IncMargin{.2em}
   
    \caption{Online TTA with adaptation gating}
    \label{alg:tta}
  
    \KwIn{Camera sequence ($\{\mathbf{V}_t\}_{t=1}^T$), IMU sequence ($\{\mathbf{I}_t\}_{t=1}^T$), frozen weight ($\mathbf{\Theta}_{f}$), adaptation weight ($\{ \mathbf{\Theta}_{a}^{k} \}_{k=0}^K$), domain distinctive feature ($\{\mathbf{d}^{k} \}_{k=0}^K$), learning rate ($\eta$)}
    \KwOut{Pose transformation sequence ($\{\hat{\mathbf{y}}^t\}_{t=1}^{T-1}$)}
    
    \For{$t := 1$ to $T-1$}
    {
        $\hat{\mathbf{y}}_f, \hat{\mathbf{y}}_i, \hat{\mathbf{d}}_t \leftarrow f(\mathbf{V}_{t \rightarrow t+1}, \mathbf{I}_{t \rightarrow t+1}, \mathbf{\Theta}_k$)\;
        $k \leftarrow \text{Match}(\hat{\mathbf{d}}_t, \mathbf{d}^k$)\tcp*[r]{Eq. \ref{eq:match}}
        \uIf{$k \neq 0$}
        {
            $\mathbf{\Theta}_{a}^k$ $\leftarrow$ $\mathbf{\Theta}_{a}^k - \eta \nabla_{\mathbf{\Theta}} \mathcal{L}_\text{TTA}(\hat{\mathbf{y}}_f, \hat{\mathbf{y}}_i)$\tcp*[r]{BatchNorm parameter update} \label{ln:BN}
        }
    }
  \end{algorithm}
\end{figure}

\csection{Online adaptation} The online TTA algorithm is delineated in Algorithm \ref{alg:tta}.
The network continuously infers and adapts when demanded by the gating signal.
For a single forward path, the network outputs two poses $\hat{\mathbf{y}}_f, \hat{\mathbf{y}}_i$ and the {\ddf}, denoted by $\hat{\mathbf{d}}$.
Domain matching algorithm is then run to identify whether the feature is in-distribution or out-of-distribution (OoD) from the source domain.
If the result is OoD, the network adapts based on test-time loss.
The network updates the BN parameters for the corresponding noise only.

\begin{figure}[t]
    \centering
    \includegraphics[width=0.8\linewidth]{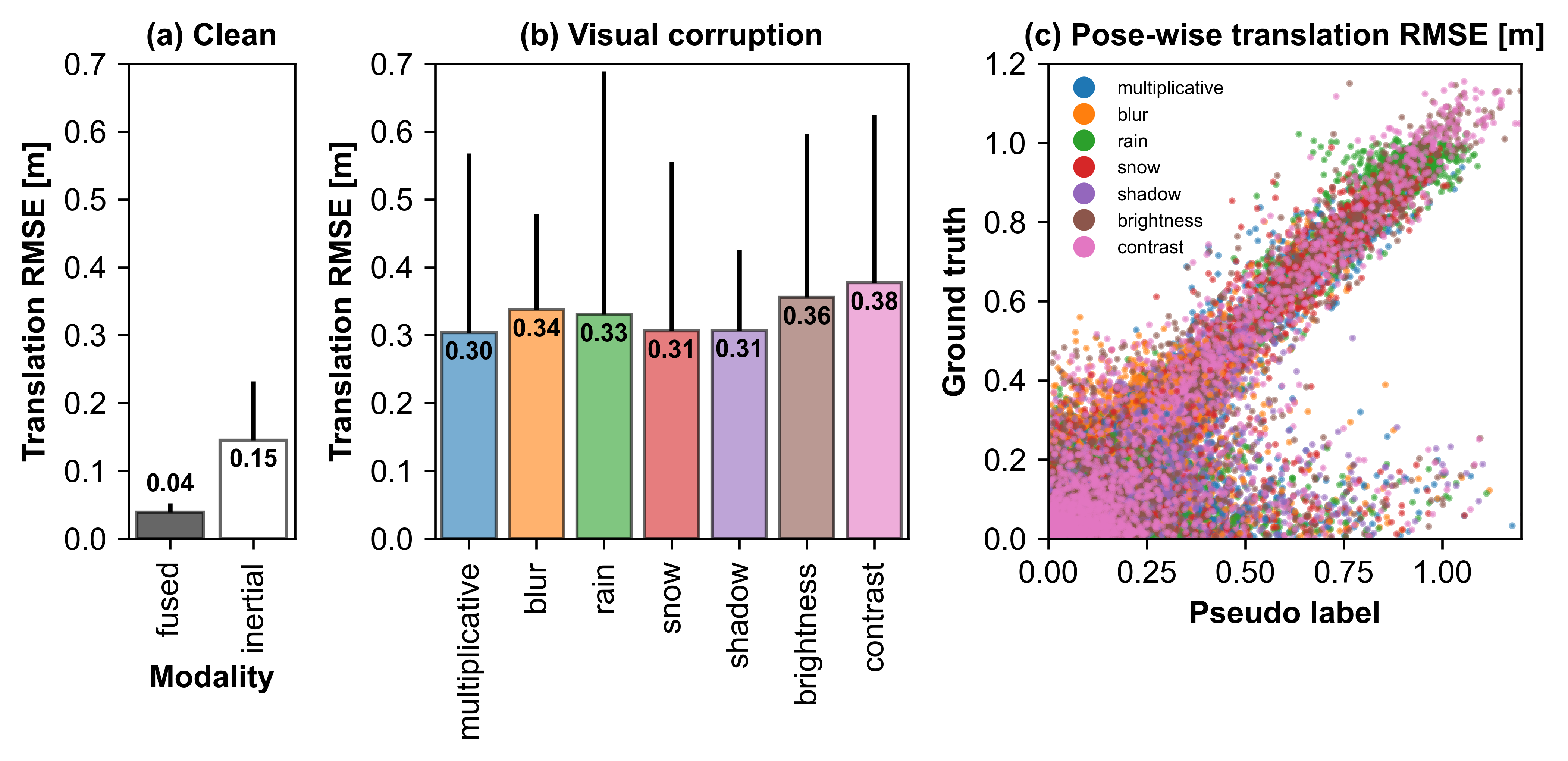}
    \caption{\textbf{Motivation for consistency loss.} (a) On a clean setting, visual feature-based inference far surpasses that of inertial. The tick represents the standard deviation. (b) Pose outputs from fused features are much affected under noisy environments. (c) A strong correlation ($r=0.86$) is shown between the relative translation error of the predicted pose against the ground truth ($x$-axis) and the inertial-inferred pseudo label ($y$-axis).}
    \label{fig:corr}
\end{figure}

\csection{Inertial-inferred pseudo label} Although the inertial-inferred pose estimates exhibit sub-par performance compared to that of vision, it is unaffected by the weather conditions.
When we simulate adversarial weather conditions on KITTI-C, we observe that the fused-feature-based poses become much more erroneous than the inertial-referred poses.
Fig. \ref{fig:corr} demonstrates a strong correlation ($r = 0.86$) between the inertial-inferred output and the ground truth.
This is obtained by evaluating pose-wise translation root mean squared error (RMSE) by comparing $\hat{\mathbf{y}}_f$ against the ground truth label $\mathbf{y}$ and the pseudo label $\hat{\mathbf{y}}_i$.

 While the loss is per batch for applying stochastic gradient descent at train time, the test-time loss function per pose corresponds to single-batch online adaptation.
\begin{equation}
\mathcal{L}_\text{TTA} = \left\lVert \hat{\mathbf{v}}_i-\hat{\mathbf{v}}_f \right\rVert ^2_2
                 + \alpha \left\lVert \hat{\mathbf{\Phi}}_i-\hat{\mathbf{\Phi}}_f \right\rVert ^2_2
\label{eq:loss_tta}
\end{equation}

\csection{Batch normalization} We dedicate a separate BN dictionary and load different sets of learnable BN parameters based on noise types.
This incurs only $0.18\%$ parameter overhead per noise type.
Solely adapting the learnable BN parameters {$\mathbf{\Theta}_a^{k}$} in the visual encoder $E_\text{visual}$ allows efficient adaptation \cite{wang2021tent}.
BN weights in the encoder are stored in and loaded from BN dictionary, whose index is decided by the domain matching algorithm, which will be explained in Section \ref{subsec:dm}.
Given a BN, $\mathbf{o}_\text{BN} =  \gamma \left( \mathbf{o} - \mu \right) / {\sigma}  + \mathbf{\beta}$, for an output feature map $\mathbf{o}$, we only update affine transformation parameters $\mathbf{\Theta}_{a} = \{ \gamma_{l,c}, \beta_{l,c} \}$ for layer $l$ and channel $c$ in $E_\text{visual}$.
The remaining parameters $\mathbf{\Theta}_{f} = \{\mathbf{\Theta}_{E_\text{v}} \setminus \{ \gamma_{l,c}, \beta_{l,c} \}, \mathbf{\Theta}_{E_\text{i}}, \mathbf{\Theta}_{D_\text{f}}, \mathbf{\Theta}_{D_\text{i}}\}$ are fixed.

\subsection{Domain matching} \label{subsec:dm}

We generate an adaptation gating signal from the domain distinctive feature ({\ddf}) $\hat{\mathbf{d}}$ to arbitrate the adaptation.
We create a {\ddf} by collecting channel-wise feature statistics of the convolution output and the activation output of the first layer \cite{gatys2016image, matsuura2020domain}.
Here, $\hat{\mathbf{d}}$ is composed of
\begin{equation}
    \hat{\mathbf{d}} = \mu(\mathbf{o}_1) \Vert \sigma(\mathbf{o}_1) \Vert \mu(\mathbf{i}_2) \Vert \sigma(\mathbf{i}_2)
    \label{eq:ddf}
\end{equation}
where \(\mathbf{o}_1\) refers to the feature map generated after the convolution in the first layer of \(E_{\text{v}}\). 
We then produce \(\mathbf{i}_2\) by applying BN and LeakyReLU to \(\mathbf{o}_1\).

The module detects a domain shift by comparing the $l^2$ norm between $\hat{\mathbf{d}}_t$ at time $t$ with {\ddf} proxies $\{\mathbf{d}^{k} \}_{k=0}^K$.
The adaptation gating signal $k_t$ is obtained by
\begin{equation}
    k_t = \mathop{\arg \min}\limits_{k \in [0,1,..,K]} \Vert\hat{\mathbf{d}}_t - {\mathbf{d}}^k\Vert_2
    \label{eq:match}
\end{equation}
which returns the index to the smallest distance.
We initialize the {\ddf} proxy by using the feature vectors of $E_{\text{v}}$ from a few images under visual corruption pre-deployment. 
We do not use source data for adaptation, while previous works such as EcoTTA \cite{song2023ecotta}, TTT \cite{liu2021ttt++}, and EATA \cite{niu2022efficient} use source data during TTA.

\section{Experiments}

\subsection{Experimental setup}
\csection{KITTI odometry dataset} \cite{geiger2012we}
Our VIO was tested with KITTI odometry dataset, which has 22 sets of driving stereo video sequences. 
Among them, Seq. $00$-$10$ contains the ground truth data and IMU readings except for Seq. $03$, and Seq. $11$-$22$ does not include the ground truth.
We follow the train/test split from previous works \cite{chen2019selective,yang2022efficient,chen2023search};
we use Seq. $00, 01, 02, 04, 06, 08$, $10$ for training and Seq. $05, 07, 10$ for testing.

\csection{EuRoC MAV dataset} \cite{burri2016euroc}
We use ten of eleven sequences for training and the remaining Seq. \textit{MH\_4\_difficult} for testing by following the train/test split in ModeSel \cite{yang2022efficient} and Hard Fusion \cite{chen2019selective}.
The grayscale images of the EuRoC MAV dataset are converted into 3-channel images.

\csection{Marulan dataset} \cite{peynot2010marulan}
We conduct \textit{real-world} domain shift experiments on the Marulan dataset to evaluate our adaptation scheme.
As intended for challenging environmental conditions, domain shifts occur naturally for conditions such as night, dust, smoke, and rain.
We use Seq. $29, 32, 33, 35, 40$ for training and Seq. $25, 36, 38, 39$ for TTA.

\csection{Vision corruption}
We apply synthetic vision corruption to the visual inputs during VIO at test time.
Such synthetic image corruption is widely adopted in prior TTA works \cite{hendrycks2019robustness, wang2021tent, wang2022continual, yuan2023robust, song2023ecotta, song2023test, boudiaf2022parameter,yuan2023robust, chen2022contrastive, zhang2023domainadaptor, chen2023improved, hatem2023point}.
This presents a significant challenge for vehicle odometry in both driving and flying scenarios, comparable to the challenges encountered in image classification or semantic segmentation.
Image manipulation was performed by using the functions provided in CIFAR-10C and ImageNet-C \cite{hendrycks2019robustness} and the Albumentation library \cite{info11020125} for additional corruptions like multiplicative, rain, snow, and shadow.

\csection{Implementation details} 
Our pre-trained model based on the source domain is implemented using PyTorch \cite{paszke2019pytorch} on a single NVIDIA Quadro RTX 6000. 
Images are resized to $512 \times 256$ during both training and adaptation.
We chose a batch size of $16$ and epochs up to $100$.
The Adam optimizer \cite{kingma2015adam} was used with a learning rate of $10^{-4}$, $\beta_1=0.9$ and $\beta_2=0.999$, and the regularization was controlled with weight decay of $5 \times 10^{-6}$.
We choose $\alpha = 100$ as the weight factor between rotational and translational vectors following \cite{chen2023search, yang2022efficient}.
We first train the conventional VIO consisting of the encoders $E_\text{visual}, E_\text{inertial}$ and the decoder $D_\text{fused}$.
Here, we employ transfer learning, but only the weights of relevant layers in $E_\text{visual}$ are initialized with that of \cite{chen2023search}.
After that, the inertial decoder $D_\text{inertial}$ is trained after freezing $E_\text{inertial}$.
The learning of $D_\text{inertial}$ was done with a batch size of $64$ for epochs up to $100$ by using inertial-inferred pose predictions $\hat{\mathbf{v}}_t^i, \hat{\mathbf{\Phi}}_t^i$ for the loss function in Eq. \ref{eq:loss_train}.
We use the same hyperparameters when performing transfer learning for the EuRoC and Marulan datasets.

\csection{Metric} Two most widely used metric to evaluate the pose estimates are based on (1) pose sequence $\{ \mathbf{p}_t \}$ and (2) camera pose transformations $\{ \mathbf{T}_{t \rightarrow t+1} \}$, which is converted to $\{ \mathbf{y}_t \} = \{ \mathbf{\Phi}_t \Vert \mathbf{v}_t \}$ for convenience. 
The RMSE error for translation and rotation vectors are calculated by $t_{rmse} = \sqrt{\frac{1}{T-1}\sum_{t=1}^{T-1} \Vert \mathbf{v}_t - \hat{\mathbf{v}}_t \Vert ^2}$ and $r_{rmse} = \sqrt{\frac{1}{T-1}\sum_{t=1}^{T-1} \Vert \mathbf{\Phi}_t - \hat{\mathbf{\Phi}}_t \Vert ^2}$. 
On the other hand, the relative translation errors ($t_{rel}$) and rotation errors ($r_{rel}$) are calculated by accounting for pose differences along 100, 200, ..., 800 meters as per \cite{geiger2012we}.

\subsection{Main results}

\begin{figure}[t]
    \begin{subfigure}[t]{.49\textwidth}  
      \centering  
      \includegraphics[width=1.0\linewidth]{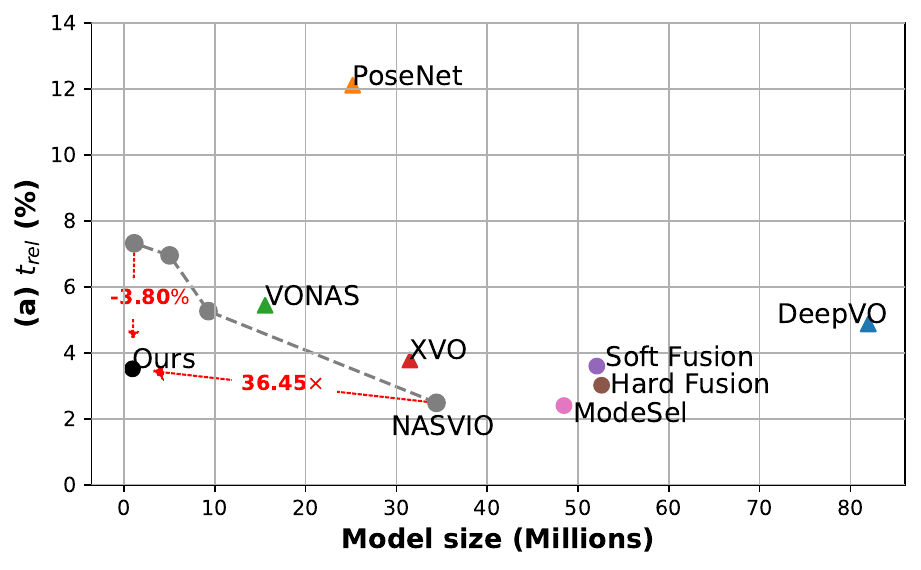}
    \end{subfigure}   
    \hfill
    \begin{subfigure}[t]{.49\textwidth}  
      \centering  
      \includegraphics[width=1.0\linewidth]{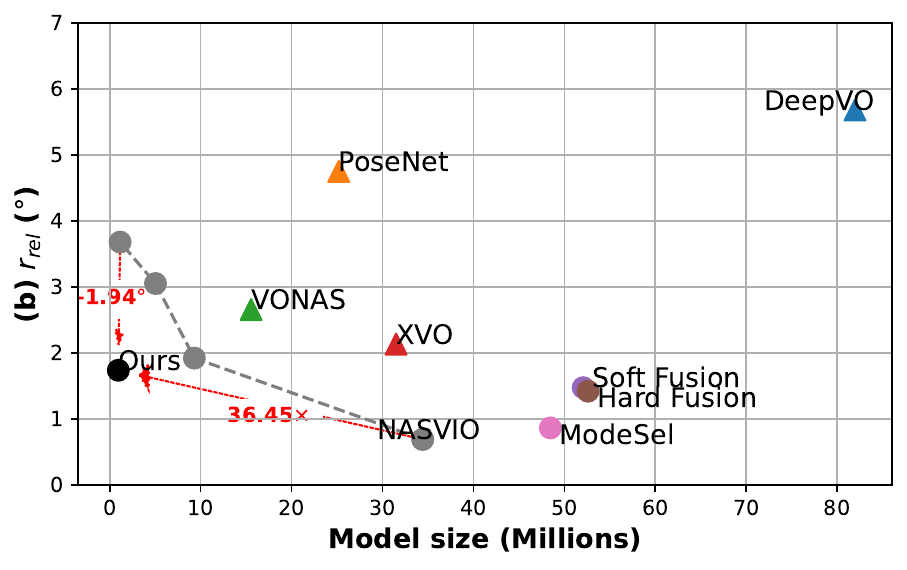}
    \end{subfigure}
  \caption{\textbf{Model size comparison} (a) relative translation error and (b) relative rotation error vs. model size comparison for supervised networks tested on KITTI Seq. $05$, $07$, and $10$. VO and VIO networks are shown as a triangle and a circle, respectively.}
  \label{fig:model_size}
\end{figure}

\begin{figure}[t]
	\centering
	\begin{subfigure}[t]{.33\textwidth}
		\includegraphics[width=1.0\textwidth]{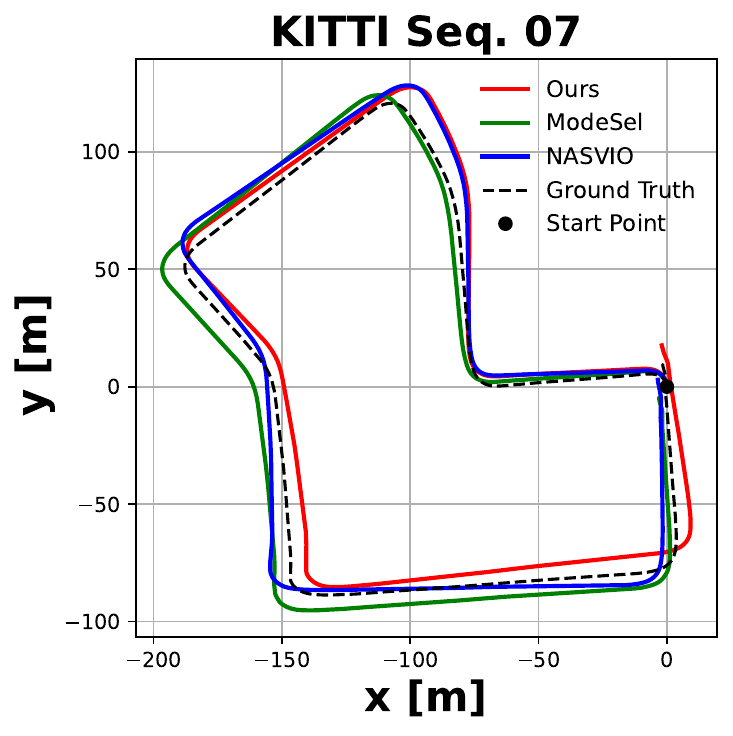}
		\captionsetup{justification=centering}
		\caption{}
	\end{subfigure}
	\begin{subfigure}[t]{.33\textwidth}
		\includegraphics[width=1.0\textwidth]{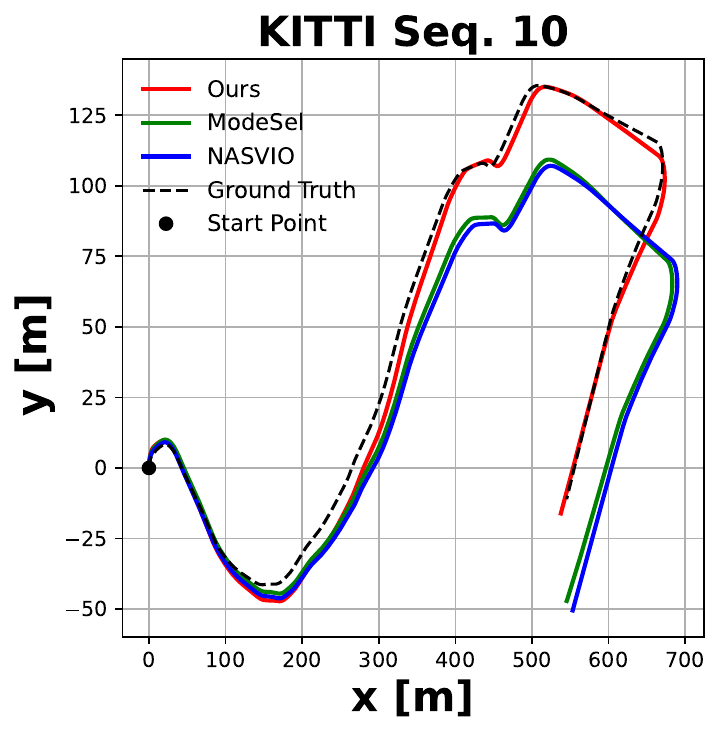}
		\captionsetup{justification=centering}
		\caption{}
	\end{subfigure}
	 
        \caption{\textbf{KITTI trajectory results} Trajectory results of our model evaluated against NASVIO \cite{chen2023search} and ModeSel \cite{yang2022efficient} on KITTI (a) Seq. $07$ and (b) $10$.}
	\label{fig:path_clean}
\end{figure}

\csection{Model compression} We compare the pose estimation error against model size for supervised networks: DeepVO \cite{wang2017deepvo}, PoseNet \cite{kendall2015posenet}, VONAS \cite{cai2020vonas}, KITTI-trained teacher model in XVO \cite{lai2023xvo}, Soft/Hard Fusion \cite{chen2019selective}, ModeSel \cite{yang2022efficient}, and NASVIO \cite{chen2023search} in Fig. \ref{fig:model_size}.
The estimation error reports are accumulated from \cite{yang2022efficient, lai2023xvo}.
Our compressed result gives $36.45 \times$ lower model size than that of the target state-of-the-art baseline, NASVIO \cite{chen2023search}, while having a minute increase in relative translation/rotation errors $\{ t_{rel}, r_{rel} \} = \{ 1.11\%, 1.05^{\circ} \}$ against the art.
For similarly-sized NASVIO maintaining the architecture, we achieve translation/rotation error reduction of $\{ 3.80\%, 1.94^{\circ} \}$.
We also compare the trajectory output of our network against the state-of-the-art in Fig. \ref{fig:path_clean}.
Our network performs comparably to others on Seq. $07$ and outperforms others on Seq. $10$.

In addition to KITTI, we report the results on the EuRoC MAV dataset \cite{burri2016euroc} in Table \ref{tab:euroc}.
Lightweight VIO is particularly relevant for aerial vehicles with limited resources.
We achieve comparable results against the state-of-the-art VIO methods \cite{yang2022efficient, chen2019selective} while decreasing the model size by orders of magnitude.

\begin{table}[t]
\scriptsize 
\centering
\begin{tabularx}{.8\textwidth}{Y|Y|Y|Y}
\toprule
 & Ours & ModeSel \cite{yang2022efficient} & Hard Fusion \cite{chen2019selective} \\
\midrule
$t_{rmse}$ [m] & 0.0282 & \textbf{0.0178} ($-$0.0104) & 0.0283 ($+$0.0001) \\
$r_{rmse}$ (\textdegree) & 0.0756 & 0.0906 ($+$0.0150) & \textbf{0.0402} ($-$0.0354) \\
Model size (M) & \textbf{0.944} & 48.454 ($\times$51.3) & 52.598 ($\times$55.7) \\
\bottomrule
\end{tabularx}
\caption{Odometry results on EuRoC \textit{MH\_4\_difficult} and model size comparison}
\label{tab:euroc}
\end{table}

\csection{TTA with stationary domain shift} We demonstrate the effectiveness of our TTA method by comparing it with networks fine-tuned with adversarial noises in Table \ref{tab:confusion}.
Except for one case, e.g., multiplicative noise, our TTA method has the best or second-best accuracy.
This case assumes stationary domain shift as in \cite{wang2021tent}.
Here, we fine-tuned the baseline model, trained initially on the noise-free source domain, by introducing the corresponding visual corruption.
For fine-tuining, we use Seq. $00, 01, 02, 04, 06, 08$, and $10$ with visual corruption for training for epochs up to twenty.
For TTA, the network is adapted on Seq. $05, 07$, and $10$ for five epochs using $\hat{\mathbf{y}}_i$ as the pseudo label.

\begin{table*}[t]
\scriptsize 
\centering
\begin{tabularx}{\textwidth}{Y|Y|Y|Y|Y|Y|Y|Y|Y|Y}
\toprule
\multicolumn{2}{c|} {\multirow{2}{*}{{\textbf{Model}}}}  & \multicolumn{8}{c}{\textbf{Average pose-wise $t_{rmse}$ [m]}} \\
\multicolumn{2}{c|}{} & \textbf{Clean} & \textbf{Multi.} & \textbf{Blur} & \textbf{Rain} & \textbf{Snow} & \textbf{Shadow} & \textbf{Bright.} & \textbf{Cont.}  \\
\midrule
\multicolumn{2}{c|}{Source}  & 0.059   & 0.154          & \underline{0.261}  & 0.176  & 0.191  & 0.203  & 0.226      & \underline{0.250}    \\
\midrule
\multirow{7}{*}{\shortstack{Fine-\\tuned\\ with\\ adver.\\ noise \\(FT)}} & Multi.         & 0.099   & \underline{0.129} & 0.394  & 0.227  & 0.372  & 0.192  & 0.299  & 0.331  \\ 

& Blur           & 0.115   & 0.176          & 0.263  & 0.193  & 0.247  & 0.184  & 0.242      & 0.261    \\ 

& Rain           & 0.289   & 0.325          & 0.372  & \textbf{0.095}  & 0.394  & 0.311  & 0.525      & 0.531     \\ 

& Snow           & 0.091   & 0.148          & 0.319  & 0.263  & \underline{0.183}  & 0.208  & 0.369      & 0.450     \\ 

& Shadow         & 0.085   & \textbf{0.112}          & 0.322  & 0.179  & 0.243  & \textbf{0.121}  & \underline{0.221}      & 0.252     \\ 

& Bright.     & 0.091   & 0.151          & 0.312  & 0.177  & 0.226  & 0.185  & 0.233      & 0.278     \\ 

& Cont.       & 0.093   & 0.150          & 0.330   & 0.197  & 0.219  & 0.184  & 0.237      & 0.273     \\

\midrule
\multicolumn{2}{c|}{\textbf{TTA (ours)}} &   -    & 0.156     & \textbf{0.230}  & \underline{0.143}  &  \textbf{0.172}  & \underline{0.155}  & \textbf{0.193}  & \textbf{0.212}  \\ 
\bottomrule
\end{tabularx}
\caption{\textbf{Comparison with networks fine-tuned with adversarial noise.} We report the average pose-wise $t_{rmse}$ results on KITTI Seq. $05$, $07$, and $10$ with noise injected throughout the series. (\textbf{Boldface} and \underline{underline} respectively indicate the best and the second-best performance.)}
\label{tab:confusion}
\end{table*}

\begin{figure}[!t]
    \begin{subfigure}[t]{.47\textwidth}
      \centering  
      \includegraphics[width=\linewidth]{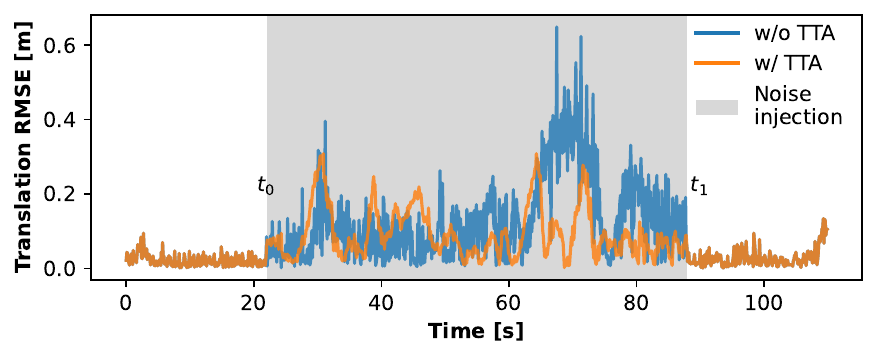}
        \caption{}   
    \end{subfigure}   
    \hfill
    \begin{subfigure}[t]{.49\textwidth}
      \centering  
      \includegraphics[width=\linewidth]{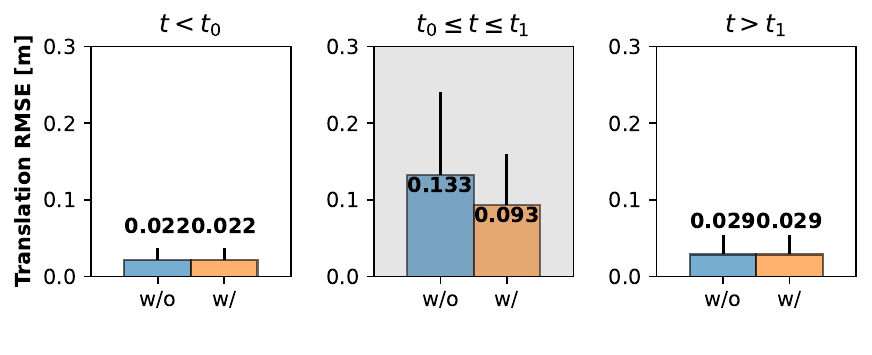}
        \caption{}
    \end{subfigure}
  \caption{\textbf{Online TTA with single domain shift} (a) Pose-wise $t_{rmse}$ and (b) avgerage pose-wise $t_{rmse}$ in the given window on KITTI Seq. $07$ with blur noise.}
  \label{fig:time_series}
\end{figure}

\csection{Online TTA with a single non-stationary domain shift}
We report VIO results for dynamically corrupted vision inputs on KITTI Seq. {\tr $07$} with and without TTA in Fig. \ref{fig:time_series}.
The sequence starts with clean images until {\tr $t_0 = 22$s}.
After $t_0$, the system instead receives blurred images, which continues until {\tr $t_1 = 88$s}.
Then, the distribution shift is removed, and the image input returns to the uncorrupted source domain.
Such a domain shift results in a pose-wise $t_{rmse}$ increase from {\tr $0.022$ m} to {\tr $0.133$ m}.
TTA reduces the error by {\tr $29.7 \%$} to {\tr $0.093$} m.
Our method also alleviates catastrophic forgetting \cite{wang2022continual} via the gating signal, which could happen if the model is continuously adapted.
Again, we mitigate the memory issue by switching the BN parameters of the visual encoder.

\begin{figure}[!t]
	\centering
	\begin{subfigure}[t]{.34\textwidth}
		\includegraphics[width=1.0\textwidth]{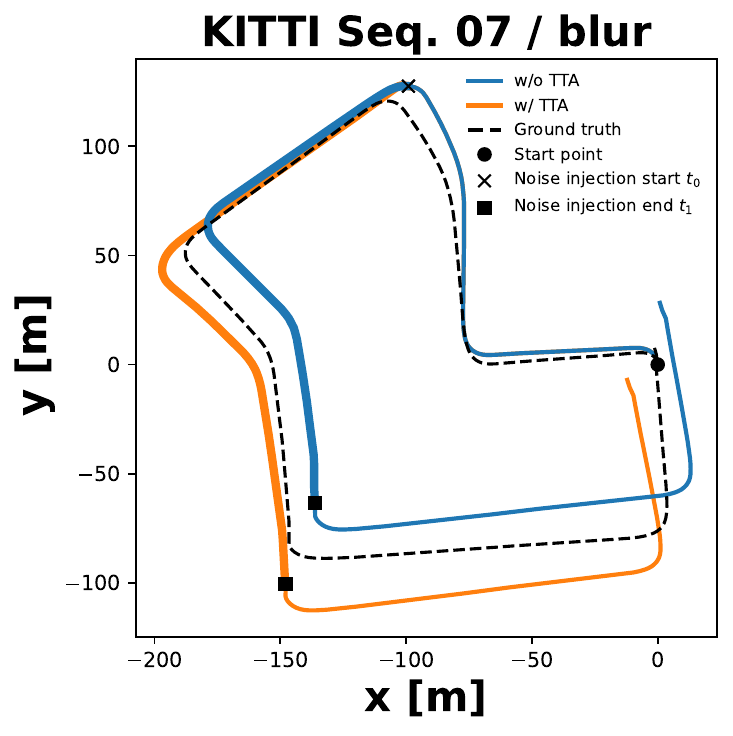}
		\captionsetup{justification=centering}
		\caption{}
	\end{subfigure}
	\begin{subfigure}[t]{.34\textwidth}
		\includegraphics[width=1.0\textwidth]{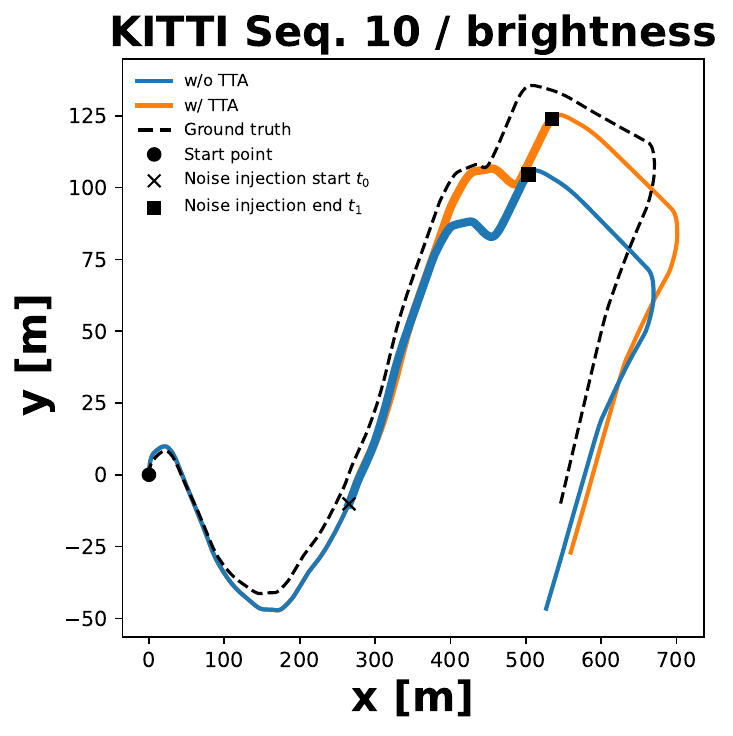}
		\captionsetup{justification=centering}
		\caption{}
	\end{subfigure}
        \caption{\textbf{Online TTA trajectory results on KITTI} Visual noise is applied to the image inputs at $t_0$ onset and is ceased at $t_1$. Our scheme adapts to such dynamic noise online in KITTI (a) Seq. $07$ with blur noise and (b) Seq. $10$ with brightness noise.}
	\label{fig:path}
\end{figure}

We illustrate the trajectory plot of the online TTA against simple inference in Fig. \ref{fig:path}.
After departing from the initial location, input distribution shifts at time $t_0$, marked with `X', due to the environmental conditions while driving.
The visual input returns to normal condition after $t_1$, represented by a square.
Due to the injected noise, the VIO network underestimates the translation vector $\hat{\mathbf{v}}$. 
Hence, the shorter distance traveled by the network performing inference without TTA.
After the noise injection ceases, the adaptation gating signal is removed, and BN weights are restored for the source domain.

\csection{Online TTA with multiple non-stationary domain shifts}
We test our domain-discriminative lightweight TTA with multiple domain shifts to simulate driving or flying scenarios experienced in the real world.
We perform vision corruptions to KITTI and EuRoC datasets with methods from  ImageNet-C \cite{hendrycks2019robustness}.
With continual TTA on KITTI, our UL-VIO achieves {\tr $18$\%} reduction in pose-wise $t_{rmse}$ on average as shown in Table \ref{tab:ctta_kitti}.
The domain-discriminative TTA governs $K$ sets of lightweight BN parameters adequately switched based on domain matching with high {\ddf} acc. of 99.6\%.
Our scheme adapts to continual domain shifts on the EuRoC dataset with similar noise settings as presented in Table \ref{tab:ctta_euroc}.
In addition, TTA performance on the Marulan dataset accompanying real-world domain shifts also demonstrates improved pose regression (Table \ref{tab:ctta_marulan}).

\begin{table}[!t]
\centering
\scriptsize 
\begin{tabular}{c|cccc|cccc|cccc|c}
\multicolumn{1}{c}{Time} & \multicolumn{12}{c}{$t \xrightarrow{\hspace*{9.3cm}}$} & \\
\toprule
\textbf{Seq.} & \multicolumn{4}{c|}{\textbf{Seq. $05$}} & \multicolumn{4}{c|}{\textbf{Seq. $07$}} & \multicolumn{4}{c|}{\textbf{Seq. $10$}} & 
\multirow{2}{*}{Avg.} \\
\textbf{Noise} & Blur & Rain & Snow & Con. & Blur & Rain & Snow & Con. & Blur & Rain & Snow & Con. & \\
\midrule
\textbf{Baseline} & 0.118 & 0.121 & \textbf{0.103} & 0.166 & 0.127 & 0.153 & 0.110 & 0.191 & 0.137 & 0.134 & \textbf{0.120} & 0.167 & 0.137 \\
\textbf{TTA} & \textbf{0.112} & \textbf{0.107} & {0.110} & \textbf{0.107} & \textbf{0.101} & \textbf{0.108} & \textbf{0.106} & \textbf{0.104} & \textbf{0.123} & \textbf{0.124} & 0.121 & \textbf{0.127} & \textbf{0.113} \\
\midrule
\textbf{\textit{ddf} acc.} & 97.9 & 100 & 100 & 100 & 98.2 & 100 & 100 & 100 & 98.8 & 100 & 100 & 100 & 99.6 \\
\bottomrule
\end{tabular}
\caption{\textbf{Continual TTA on KITTI} Average pose-wise $t_{rmse}$ and {\ddf} accuracy ($K=4$) measured on KITTI Seq. $05$, $07$, $10$ with cyclical vision corruptions.}
\label{tab:ctta_kitti}
\end{table}

\begin{table}[!t]
\centering
\scriptsize 
\begin{tabularx}{.65\textwidth}{c|Y|Y|Y|Y}
\multicolumn{1}{c}{Time} & \multicolumn{3}{c}{$t \xrightarrow{\hspace*{4cm}}$} & \\
\toprule
\textbf{Noise} & {Blur} & {Bright.} & {Contrast} & {Avg.} \\
\midrule
\textbf{Baseline} & 0.0255 & 0.0256 & 0.0276 & 0.0262 \\
\textbf{TTA} & \textbf{0.0253} & \textbf{0.0254} & \textbf{0.0254} & \textbf{0.0254} \\
\midrule
\textbf{\textit{ddf} acc. (\%)} & 95.6 & 100.0 & 100.0 & 98.5 \\
\bottomrule
\end{tabularx}
\caption{\textbf{Continual TTA on EuRoC} Average pose-wise $t_{rmse}$ and {\ddf} accuracy ($K=3$) measured on EuRoC \textit{MH\_4\_difficult} with continual vision corruptions.}
\label{tab:ctta_euroc}
\end{table}

\begin{table}[!t]
\centering
\scriptsize 
\begin{tabularx}{.75\textwidth}{c|Y|Y|Y|Y|Y}
\multicolumn{1}{c}{Time} & \multicolumn{4}{c}{$t \xrightarrow{\hspace*{5cm}}$} & \\
\toprule
\textbf{Cont. Seq.} & $25$-Night & $36$-Dust & $38$-Smoke & $39$-Rain & {Avg.} \\
\midrule
\textbf{Baseline} & 0.244 & {0.230} & \textbf{0.228} & 0.248 & 0.237 \\
\textbf{TTA} & \textbf{0.227} & \textbf{0.227} & {0.233} & \textbf{0.241} & \textbf{0.232} \\
\midrule
\textbf{\textit{ddf} acc. (\%)}  & 98.6 & 90.5 & 100.0 & 100.0 & 97.3 \\
\bottomrule
\end{tabularx}
\caption{\textbf{Continual TTA on Marulan} Average pose-wise $t_{rmse}$ and {\ddf} accuracy ($K=4$) measured on Marulan with varying environmental noise innate to the dataset.}
\label{tab:ctta_marulan}
\end{table}

\begin{figure}[!t]
	\centering
	\begin{subfigure}[t]{.3\textwidth}
		\includegraphics[width=1.0\textwidth]{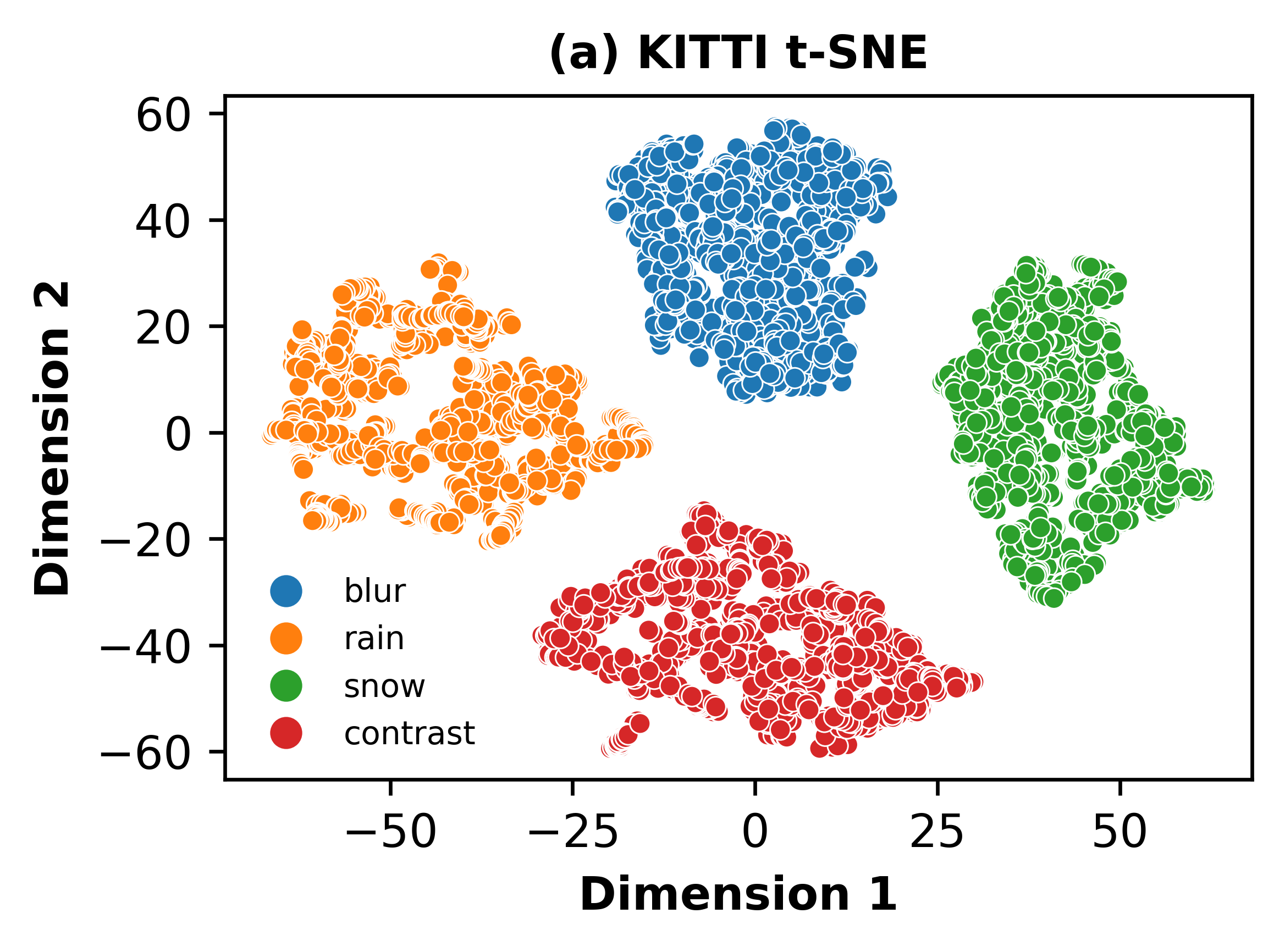}
		\captionsetup{justification=centering}
	\end{subfigure}
	\begin{subfigure}[t]{.3\textwidth}
		\includegraphics[width=1.0\textwidth]{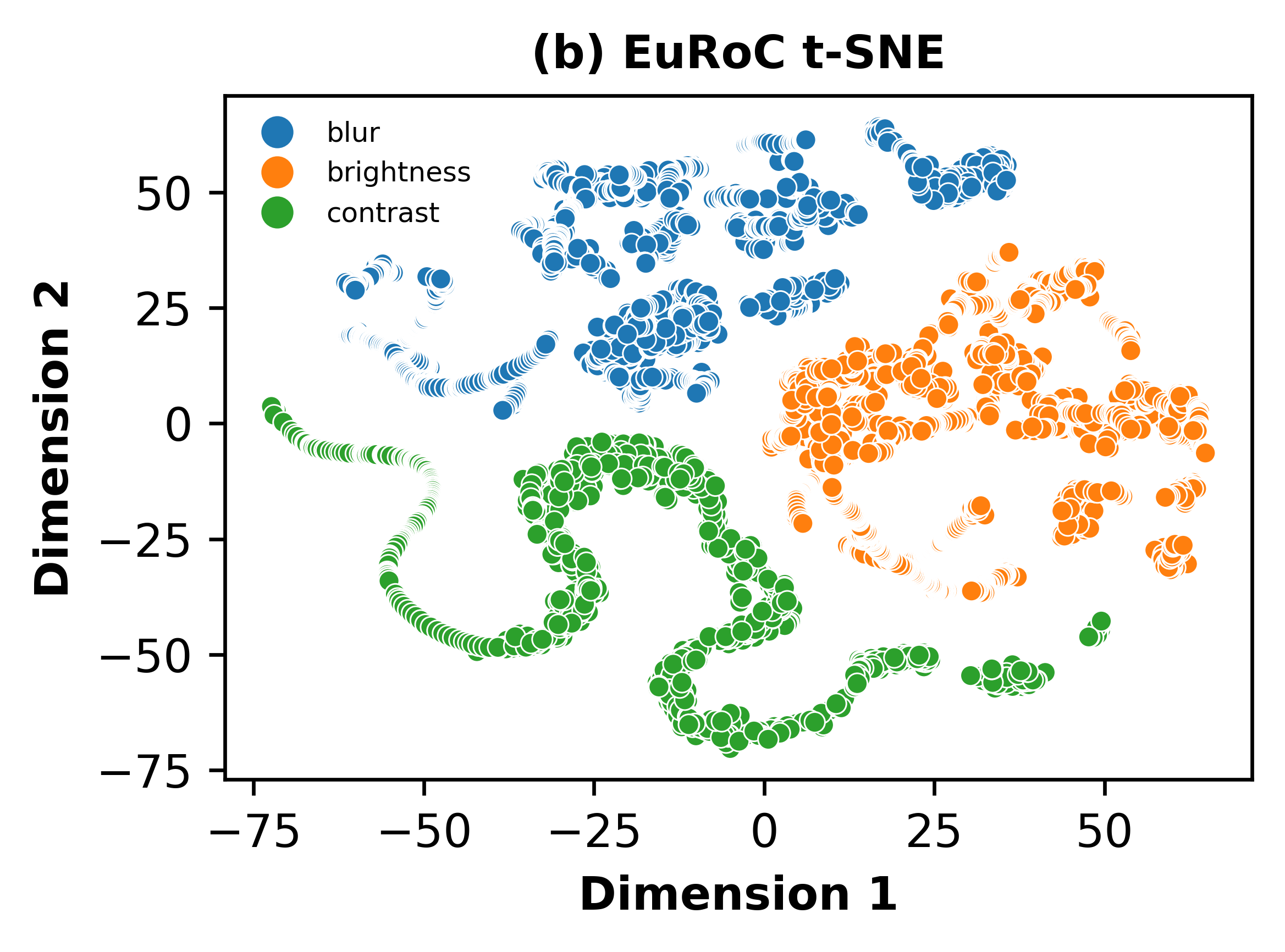}
		\captionsetup{justification=centering}
	\end{subfigure}
	\begin{subfigure}[t]{.3\textwidth}
		\includegraphics[width=1.0\textwidth]{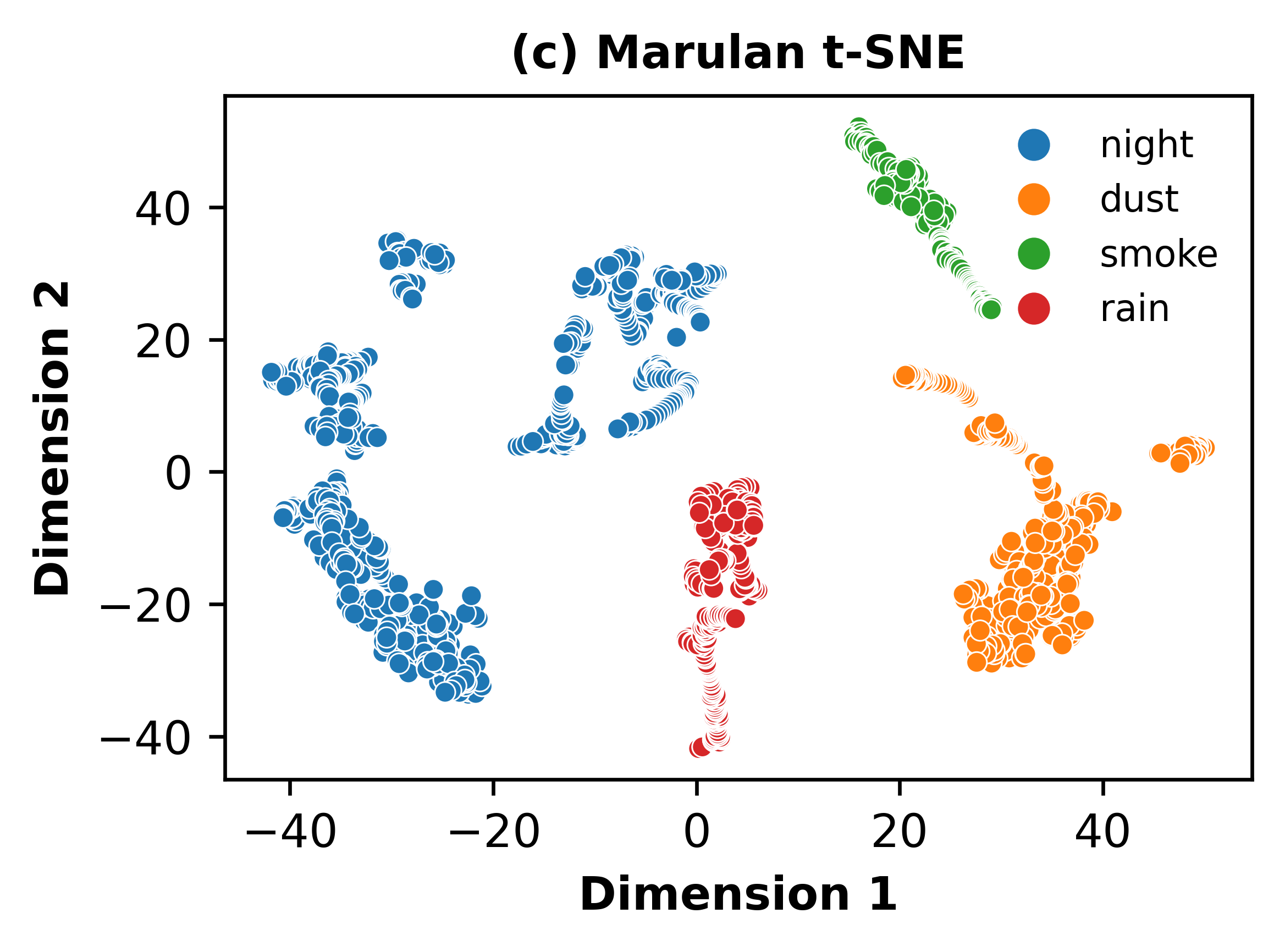}
		\captionsetup{justification=centering}
	\end{subfigure}
	 
        \caption{\textbf{t-SNE visualized domain-distinctive  features} \textit{ddf}s are well separated in cases of (a) KITTI, (b) EuRoC, and (c) Marulan.}
	\label{fig:tsne}
\end{figure}

\csection{Domain matching} We demonstrate the effectiveness of our domain matching module.
Well spaced out {\ddf}s visualized with t-SNE \cite{van2008visualizing} in Fig. \ref{fig:tsne} support high accuracy in $K$-way domain detection in Table \ref{tab:ctta_kitti}, \ref{tab:ctta_euroc}, and \ref{tab:ctta_marulan}.
We also highlight that the \textit{latency} required for domain matching is a single timestamp ($t \rightarrow t+1 $) since the feedforward $E_\text{visual}$ is memoryless, and pose regression is non-sequential and independent among consecutive poses.

\csection{Limitations} This work has a few limitations. Firstly, it relies on IMU readings, which may not always be accurate or available. Secondly, the finite dictionary size for domain-matching linearly increases with the number of domain shifts.

\section{Conclusion}
In this work, we propose UL-VIO, an ultra-lightweight VIO network capable of efficient adaptation for autonomous platforms.
We achieve a network with $<1$M parameter size through model compression, delivering $36 \times$ smaller size with a minute hit ($1\%$) on pose accuracy compared to the previous state-of-the-art.
Our lightweight model also supports resource-efficient test-time adaptation to the changing environments on the fly through visual-inertial consistency.
The proposed scheme tested on the KITTI dataset can reduce translation RMSE by up to {\tr $45$\%} depending on the noise type ({\tr $18$\%} on average) while incurring only $0.18\%$ parameter re-write overhead as it updates only the BatchNorm parameters.
We confirm the effectiveness of our lightweight adaptation scheme across various dynamic environments.

\csection{Acknowledgements:} This work was supported in part by COGNISENSE, one of seven centers in JUMP 2.0, a Semiconductor Research Corporation (SRC) program sponsored by DARPA. The work of SY Chun was supported by the Institute of Information \& communications Technology Planning \& Evaluation (IITP) grant funded by the Korea government (MSIT) [NO.RS-2021-II211343, Artificial Intelligence Graduate School Program (Seoul National University)] and by the National Research Foundation of Korea (NRF) grant funded by the Korea government (MSIT) [No. NRF-2022M3C1A309202211]


%
%
\bibliographystyle{splncs04}
\bibliography{main}
\end{document}